\documentclass[a4paper,conference,twoside]{IEEEtran}

\usepackage{cite}
\usepackage{booktabs}
\usepackage{comment}
\usepackage{graphicx}
\usepackage[caption=false,labelformat=simple]{subfig}
\usepackage{multirow}
\usepackage{flushend}

\usepackage[vmargin=0.2cm,hmargin=1cm,head=16pt,includeheadfoot]{geometry}
\usepackage{fancyhdr}
\newcommand{\etal}{\textit{et al. }}

\ifCLASSINFOpdf
\else
\fi

\usepackage{amsmath}
\usepackage{mathtools}
\usepackage{xcolor}
\DeclarePairedDelimiter\floor{\lfloor}{\rfloor}

\usepackage{array}
\usepackage{amsfonts}

\begin{document}
\title{GuCNet: A Guided Clustering-based Network for Improved Classification}

\author{\IEEEauthorblockN{Ushasi Chaudhuri, Syomantak Chaudhuri and Subhasis Chaudhuri}
\IEEEauthorblockA{
Indian Institute of Technology Bombay\\
Mumbai, India\\
Email: \{ushasi, syomantak, sc\}@iitb.ac.in}}

%\and
%\IEEEauthorblockN{Suomantak Chaudhuri}
%\IEEEauthorblockA{Indian Institute of %Technology Bombay\\
%Mumbai, India\\
%Email: syomantak@iitb.ac.in}
%\and
%\IEEEauthorblockN{Subhasis Chaudhuri}
%\IEEEauthorblockA{Indian Institute of %Technology Bombay\\
%Mumbai, India\\
%Email: sc@ee.iitb.ac.in}}


%Telephone: (800) 555--1212, Fax: (888) 555--1212}


%\markboth{In ICPR 2020}

\maketitle

\begin{abstract}
We deal with the problem of semantic classification of challenging and highly-cluttered dataset. We present a novel, and yet a very simple classification technique by leveraging the ease of classifiability of any existing well separable dataset for guidance. Since the guide dataset which may or may not have any semantic relationship with the experimental dataset, forms well separable clusters in the feature set, the proposed network tries to embed class-wise features of the challenging dataset to those distinct clusters of the guide set, making them more separable. Depending on the availability, we propose two types of guide sets: one using texture (image) guides and another using prototype vectors representing cluster centers. Experimental results obtained on the challenging benchmark RSSCN, LSUN, and TU-Berlin datasets establish the efficacy of the proposed method as we outperform the existing state-of-the-art techniques by a considerable margin.
\end{abstract}


% \fi
\IEEEpeerreviewmaketitle

\section{Introduction}
\label{sec:intro}
%How classification/clustering is important
We consider the classification problem that requires us to extract relevant features from patterns and project it onto an embedding space where the representations of each class of patterns are uniquely distinguishable. 

The most common way of tackling the problem is to map patterns to a latent space where we expect the mapping for a given semantic class of images to be compact and well separable among different classes. To ensure that, we adopt various types of measures like keeping margins in cost functions so that the representation of instances of different types are pushed apart in the feature space. A disadvantage of this method is that the corresponding embeddings of features for a given class of objects need not be compact in the feature space, particularly so during the testing phase. Some notable works that can be seen using this kind of an approach are \cite{Pasolli2016,Pernkopf2012,siam2019}.

Another common way to address the classification problem is by taking help from some additional information, like using the word2vec projections~\cite{le2014distributed} of the class labels to help in the classification task. This kind of approach has been used in \cite{akata2015label}. The approach of adding word2vec vectors from a semantic space and merging it with the features extracted from the visual space could prove to be non-beneficial as they are from completely different spaces. It is not guaranteed that there might be an one-to-one correspondence between the semantic vector and the visual features. Although quite a few papers show the utility of it in terms of enhanced classifiability~\cite{akata2015label,dutta2019semantically}, a simple exercise of randomly permuting (or rotating all these vectors by an arbitrary rotation matrix) the word2vec representations of semantics before merging yields very little difference in the classifier performance. Notwithstanding the above, if we were allowed to redesign the vector representations appropriately, how should we do it systematically to improve the classification accuracy? We plan to address this issue in this paper.

\begin{figure}
    \centering
    \includegraphics[width=0.9\linewidth]{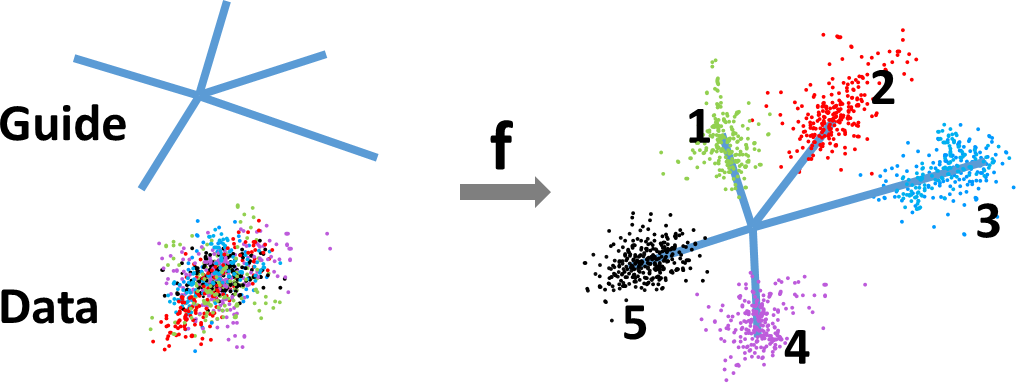}
    \caption{An illustrative figure showing the overall concept of guided clustering of data instances in the feature space by leveraging the separability of the guide data. Here the vectors represent the cluster centers of the guide data. Here \textbf{f} is the mapping that allows data to be as per guide set.}
    \label{fig:short}
\end{figure}

There is also an additional problem regarding the actual meaning of the two different spaces. The word2vec embedding for a class is unique, however, the visual samples could have numerous different embeddings depending on the orientation, scale, translation, and illumination of the image. Since the word2vec embedding is fixed, it could be difficult to map these varied set of visual features to a single semantic vector. In order to tackle this, researchers attempt to preserve the semantic topology in both the spaces using topology graphs~\cite{akata2015label,dutta2019semantically,gune2018structure}. However, the topology itself is a function of the learned embedding and is not known apriori.

%short literature survey
There are several types of standard datasets in computer applications, which are commonly used for experiments. For classification purposes, researchers may use indoor scenes, aerial scenes, sketch data, character recognition data, among others. Existing  work in the literature mostly tackles outdoor scene images using remote sensing techniques. In \cite{ali2019modeling}, the authors define a geometric model for the spatial information to effect rotation invariance and achieve the state-of-the-art (SOTA) results. The problem of outdoor scene image classification has also been attempted by \cite{krizhevsky2012imagenet,xia2017aid,zou2015deep,zafar2018novel}, and they relate to our work. Similarly, for indoor scene classification, there is a limited volume of works that have achieved the equivalent SOTA performance as it is usually considered to be a very challenging problem. The authors in \cite{wang2017knowledge} used multi-resolution CNNs for reducing the classification ambiguity in images by providing supplemental knowledge from the confusion matrix. Likewise,  \cite{wang2019learning} uses a generative adversarial network-based learning technique to improve the discriminative property of the learnt features. Since an image is quite rich in the feature space, researchers have also attempted to design appropriate networks to derive a good feature for the classification of sketch images. Since sketches are meant to be a very minimal and often incomplete description of objects, it is expected that sketch classification would be quite difficult. Most techniques in the literature use variational auto-encoders and along with adversarial training techniques for sketch classification. Some of the existing works in literature include ~\cite{krizhevsky2012imagenet,eitz2012humans,he2016deep,prabhu2018hybrid,zhang2019learning}. 
%We choose diverse vision datasets to highlight the %efficacy of our proposed classification technique.   

%Our contribution
In contrast to the methods used in literature, we depend on two simple but well-accepted facts in designing our classifier.  Firstly, the classifiability of patterns depends on the separability of features. Secondly, it is now believed that a deep neural network is quite adept in learning various different types of mapping functions. Let us now suppose that a specific classification problem is extremely challenging and the features learnt from the training data remain quite inseparable in the embedding space. But there may exist a completely different dataset where the classifier learning is very easy and the corresponding features are well separable. Examples of such well separable datasets include MNIST~\cite{lecun1998gradient} or the CIFAR~\cite{krizhevsky2009learning} dataset. Since almost any mapping can be reliably learnt from the exemplars, why not we then find an appropriate network that will map the features of the challenging dataset to those of the MNIST or CIFAR features class-wise? By doing so we expect the challenging dataset to be classified much more accurately. A further twist to this requires that instead of pre-learning MNIST features first and the mapped features of the challenging dataset, we should try to do an end-to-end learning of  both features to improve the accuracy further. Hence,  whenever we expect to encounter a feature space that is not well separable for some reason, we propose to use any other well-classifiable dataset with the same number of classes as a guide data to design a classifier for improved accuracy. This guide data may or may not at all be semantically related to the experimental dataset to be worked on. We leverage specifically this cluster-wise dissociating capability of the well-classifiable guide data for designing the new classifier for the current dataset. The concept has been illustrated in Fig.~\ref{fig:short}. Since the semantic labels, if any, of the guide data have no correlation with the semantic labels of the actual data to be classified, we call our method as a guided clustering-based network as only the clusterability property of the guide data is of importance and not their semantic labels.

The guided clustering can be done in two ways. If we have an available well classifiable dataset that has at least the same number of clusters as the dataset that is to be classified, then we can use this available data as well-separable textures to guide the classification process. In case we do not have such a separable texture data, we can actually construct a vector space representation of well separable cluster centers (as opposed to word2vec representations) to guide the neural network. One possible way of doing this can be by setting each guide class center as a vertex of a hyper-cube of an appropriate dimension. We demonstrate through ample experiments the superior performance of the proposed GuCNet architecture. We summarize the major contributions in this paper as:

\begin{itemize}
    \item We propose a guided clustering framework for classifying highly cluttered patterns in the feature space.
    \item We demonstrate that leveraging on the well classifiability of a guide data, an improved classifier can be learned for complex patterns. 
    \item In the absence of a separable texture guide, we propose an alternate prototype-based guided clustering mechanism which has been shown to perform equally well.
    \item We perform experiments on diverse benchmark computer vision datasets and show a significant boost in performance using our method.
\end{itemize}

% \section{Related Works}
% \label{sec:literature}

\section{Guided Network Learning}
\label{sec:gc}
\noindent \textbf{Preliminaries:}  Let us denote the two datasets corresponding to the challenging and the relatively simpler (from the classification point of view) dataset as  $\mathcal{X}$ and $\mathcal{Y}$, respectively. Let $\mathcal{F}_y : \mathcal{Y} \mapsto \mathcal{Y}_c$, where $\mathcal{F}_y$ is any suitable neural network that maps the raw data $\mathcal{Y}$ to a nicely separable feature space $\mathcal{Y}_c$. We aim to find a mapping $\mathcal{F}_x$ such that $\mathcal{F}_x : \mathcal{X} \mapsto \mathcal{Y}_c$ so that if $\mathcal{Y}$ is well classifiable, so would be $\mathcal{X}$ that we are interested in classifying. It is to be noted that the data stream $\mathcal{X}$ need to have semantic labels from $\mathcal{Z} \in [1,2,3....C]$, where $C$ is the number of classes for training purposes. While the guide data stream $\mathcal{Y}$ does not require specific semantic labels $\mathcal{Z}$, it does require the knowledge of the entire data stream $\mathcal{Y}$ to be split into $C$ number of clusters, following which each cluster is arbitrarily assigned a unique label in $\mathcal{Z}$. Using these, we create triads for training the model as $ \{(x_i,y_i,z_i)\}$, where $x_i \in \mathcal{X}$, $y_i \in \mathcal{Y}$, and $z_i \in \mathcal{Z}$. Our main objective is to create a unified latent feature space using the classifier of $\mathcal{Y}$, so that the embedding of $\mathcal{X}$ into this latent space is semantically discriminative.

\begin{figure*}
    \centering
    \includegraphics[width=\linewidth]{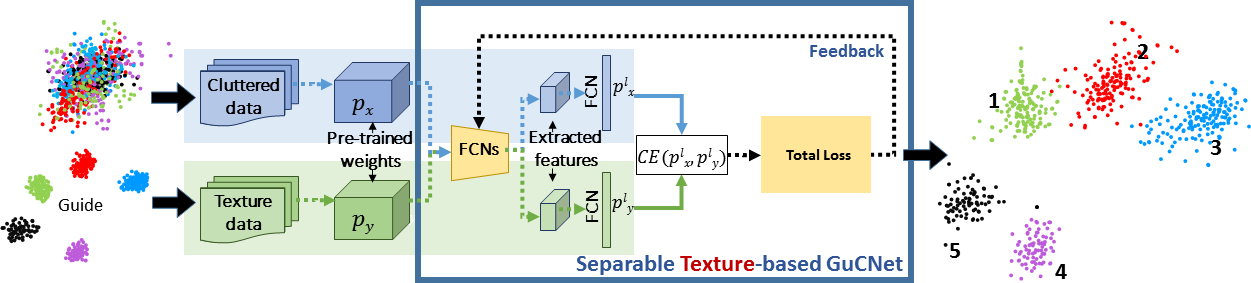}
    \caption{The overall pipeline of the proposed separable texture guide based GuCNet. The network uses the separable clusters of the texture guide and minimizes the cross-entropy of the features from the data as well as the texture guide.}
    \label{fig:Tbigblock}
\end{figure*}

We propose two possible types of frameworks for the proposed algorithm. In the first framework when examples from a similar modality of data (say, both $\mathcal{X}$ and $\mathcal{Y}$ are images of different objects) are available, we use such texture samples as the guide data $\mathcal{Y}$. Note that the textures (say, optical characters) can be totally uncorrelated with the actual data $\mathcal{X}$ (say, animal pictures). For the second framework when not enough good examples are available for guide set $\mathcal{Y}$, we use a single vector-based guide (called {\em prototype} here) for each cluster to help improving the classifier for concerned dataset $\mathcal{X}$. We explain both these frameworks in detail in the following sub-sections.

\subsection{Separable Texture-based Guiding}
\label{sec:tgc}

Under this setup, we aim to make the cluttered dataset more separable in the visual feature space. In order to do so, we take the help of another well-classifiable visual texture data as the guide $\mathcal{Y}$. For this purpose, we randomly select any existing dataset, like MNIST, as  $\mathcal{Y}$ for which most researchers have reported achieving a very high discriminability in the visual feature space using standard neural network architecture.

Using the help of this texture-guide, we learn a unified feature representation space, wherein the embeddings of both the data  $\mathcal{X}$ and  $\mathcal{Y}$ overlap class-wise (see Fig.~\ref{fig:short}). Since  $\mathcal{Y}$ is well separable, we expect  $\mathcal{X}$ to be well separable also, provided we are able to learn a good embedding. The guide data instance is contained in $y_i \in \mathbb{R} ^{N \times K}$, where $K$ is the dimension of the features extracted from the guide data and $N$ is the number of data samples in $\mathcal{Y}$. The separable texture-based guide helps us in making the unified space class-wise as discriminative as possible while bridging the domain gap between the texture data and the concerned data. %Therefore in \textbf{U}, both the projections of $\mathcal{X}$ and  $\mathcal{Y}$, which are \textbf{$U_x$} and \textbf{$U_y$}, respectively, have highly similar features and are semantically discriminative. 
To realize the unified latent feature space, we propose a simple two-stage training setup. 

\noindent \textbf{Stage I:} We find the initial level features from the data using a pre-trained network. We perform transfer-learning of knowledge from a pre-trained network (trained on the Imagenet dataset) and fine-tune the network further to align the visual features with the given dataset  $\mathcal{X}$. Similarly, we extract the initial level features from a fine-tuned pre-trained network for the texture guide data  $\mathcal{Y}$. Let us denote these extracted features as $p_x$ and $p_y$ for the cluttered-image and texture guide data, respectively. It may be noted that $p_x$ and $p_y$ could be of different dimensions.

\noindent \textbf{Stage II:} Now we need to find $\mathcal{F}_x : p_x^l \mapsto p_y^l $, where $p^l$ represents the latent layer representation of feature $p$, with superscript denoting either the cluttered or guide data. We take $p_x$ and $p_y$ as the inputs to the next stage. In this stage we place a series of fully-connected network (FCN) to extract the features of the cluttered-data and the texture data as $p_x^l$ and $p_y^l$, respectively. %The CNNs are followed by a batch-normalization layer and 
We induce non-linearity into the network by using appropriate activation functions. We then further train the network to generate the unified latent feature space from $p_x^l$ and $p_y^l$ by minimizing the overall loss function. We define the loss function ($\mathcal{L}$) as %the sum of 
the cross-entropy loss ($\mathcal{L_{\text{CE}}}$) % and the matching loss ($\mathcal{L_{\text{ml}}}$) 
in the latent layer. We define the loss function in sub-section~\ref{subsec:objective}. It may be noted that the purpose of this paper is not to design a novel neural network architecture to solve the basic classification problem, rather we want to demonstrate the utility of a good guide data to improve the classification accuracy. Hence we use a very simple neural network architecture in this paper. The details of the proposed method are illustrated in Fig.~\ref{fig:Tbigblock}.

\subsection{Separable Prototype-based Guiding}
\label{sec:pgc}
\begin{figure*}[!h]
    \centering
    \includegraphics[width=\linewidth]{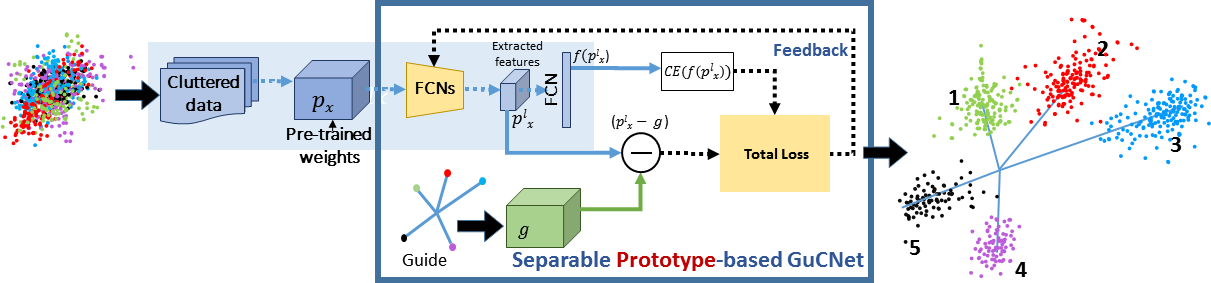}
    \caption{The overall pipeline of the proposed separable prototype guided GuCNet. The network uses the cluster centers defined by the prototype guide  $g$ to compute the matching loss and minimizes the cross-entropy of the features from the data. In the figure $f(p^l_x)$ stands for the last layer output of the classifier having a dimension of $C$ number of classes.}
    \label{fig:Pbigblock}
\end{figure*}
Under this setup, we aim to make the given cluttered data $\mathcal{X}$ more separable by cooking up a simple prototype-based guide when one cannot find a suitable texture data as the guide. For this purpose, we simply choose $K$ dimensional vectors (called {\em prototypes}), where $K \ge C$ and $C$ is the number of classes in the cluttered dataset. In effect, instead of specifying the textural features for each class as the guide data, we simply provide the locations of the cluster centers for $C$ classes to which one should try to map the latent layer features of different classes of the cluttered dataset.

At this point, one may ask how does one select the class prototypes. One may think of using the {\em word2vec} representations of class semantics as the prototypes. However, this is not a good choice as the distances among the corresponding class prototypes will be quite uneven, resulting in poor classification accuracy. Since the prototypes should be well separable, in this study we use $C$ number of unit vectors along the axes of a $K$ -dimensional Cartesian coordinate system as prototypes. Thus we have a Hamming distance of $2 \times \floor*{\frac{K}{C}}$ between any two class prototypes. It may also be noted that the assignment of a particular prototype to a specific class in the given dataset $\mathcal{X}$ has nothing to do with the class semantics.

We try to learn a unified feature representation space using this separable prototype-based guide. The guide instances are contained in $y_i \in \mathbb{R}^{C \times K}$. Similar to the previous setup, we perform a two-stage training protocol to realize the proposed prototype-based GuCNet.

\noindent \textbf{Stage I:} We find the initial level features from the data using a pre-trained network, similar to the previous case, and get $p_x$ for the cluttered data input. However, unlike in the previous case, we do not need to pre-train the guide data here as they correspond to well-separated user assigned cluster centers for individual classes in the feature space. We refer to this guide prototypes as $g$.

\noindent \textbf{Stage II:} We use $p_x$ as input to the subsequent FCN network. We refer to the output of the convoluted features from $p_x$ as $p_x^l$. To synthesize the unified semantic space, we take samples from $p_{x_c}$ and $g_c$ of class $c$ and treat them as the same semantic label $c$. The network minimizes the cross-entropy of the visual data stream and the matching loss between $p_x^l$ and $g$. Again, to minimize the matching loss, we ensure that the dimensions of $p_x^l$ and $g$ are the same. The details of the proposed method is illustrated in Fig.~\ref{fig:Pbigblock}.
%and $p_y$ for the cluttered-image and texture data, respectively. $p_x$ and $p_y$ could be of different dimensions.

\subsection{Objective Function}\label{subsec:objective}
As mentioned above, the overall objective function involves the cross-entropy loss and, additionally, the matching loss between $p_x^l$ and $g$ in case of prototype based guidance. 

\noindent \textbf{Matching loss ($\mathcal{L_\text{ml}}$):} We define the matching loss in equation~\ref{eq:matching} %$\mathcal{L_\text{ml}} = \mid~p_x^l - g~\mid$ 
for the prototype-based guides only. This cannot be defined for the texture-based guides as there is no correspondence between the elements of set $\mathcal{X}$ with those of $\mathcal{Y}$. This loss helps in bringing the guide and the cluttered data together in the feature space. This loss also helps in explicitly achieving domain invariance in the unified feature space by bridging the domain gap between the two features, as the prototype domain is very different from the visual features.

  \begin{equation}\label{eq:matching}
      \mathcal{L_\text{ml}} = \mid p_x^l - g \mid.
  \end{equation}

\noindent \textbf{Cross-Entropy loss ($\mathcal{L_\text{CE}}$):} The key loss function used in the proposed system is the cross-entropy loss. This helps in making the unified space semantically discriminative. This loss ensures that the inter-class distances among different data samples are pushed apart, while reducing the inter-modal distances between $p_x^l$ and $p_y^l$. We feed samples of class-$c$ of both $\mathcal{X}$ and  $\mathcal{Y}$ together as the same class label in the unified space while minimizing the $\mathcal{L_\text{CE}}$ loss, defined in equation~\ref{eq:ce}.%CE ($p_x^l$, $p_y^l$).

\begin{equation}\label{eq:ce}
    \mathcal{L_\text{CE}} = \text{CE} (p_x^l, p_y^l).
\end{equation}

In case of prototype based GuCNet, $\mathcal{L_\text{CE}}$ is defined as CE($f(p^l_x)$), where $f(p^l_x)$ is the output of the softmax layer and $p^l_x$ is the layer just before that. The overall objective function for the texture-based guide is only the cross-entropy loss $\mathcal{L_\text{CE}}$, while for the prototype-based it is the sum of the matching loss and the cross-entropy loss, given by %$\mathcal{L} = \mathcal{L_\text{CE}} +  \alpha \mathcal{L_\text{ml}}$.  
equation~\ref{eq:objective}. 

 \begin{equation}\label{eq:objective}
     \mathcal{L} = \mathcal{L_\text{CE}} +  \alpha \mathcal{L_\text{ml}}.
 \end{equation}
 
Here $\alpha (< 1.0)$ is an appropriate weight that encourages the minimization of the cross-entropy loss more than the matching loss. This is done in order to avoid learning up a trivial solution due to the difference operation in the matching loss. We solve the optimization problem by alternately minimizing each loss individually while keeping the other one frozen. By doing so, we transform back the non-convex optimization to a convex optimization problem for that particular loss. %We use the standard Adam optimizer for minimizing the overall objective function using a stochastic mini-batch gradient-descent approach.

\section{Experiments}
\label{sec:exp}
In this section, we discuss the chosen datasets, model architecture, and the training and evaluation protocols used in our framework. 

\begin{figure}%[!h]
\centering
{\includegraphics[height=2.05cm]{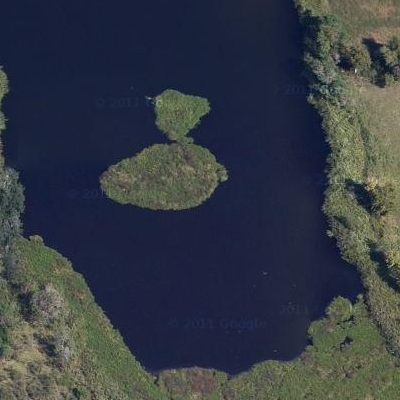}} 
{\includegraphics[height=2.05cm]{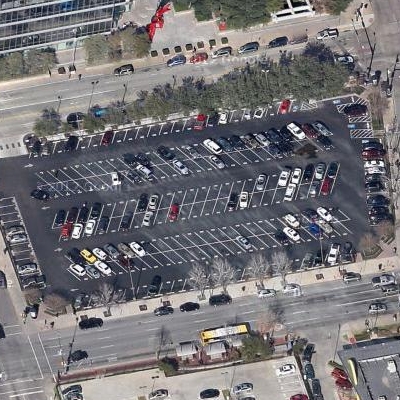}} 
{\includegraphics[height=2.05cm]{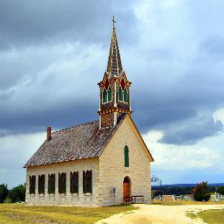}}
{\includegraphics[height=2.05cm]{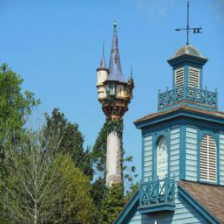}} \\
{\includegraphics[height=2.05cm]{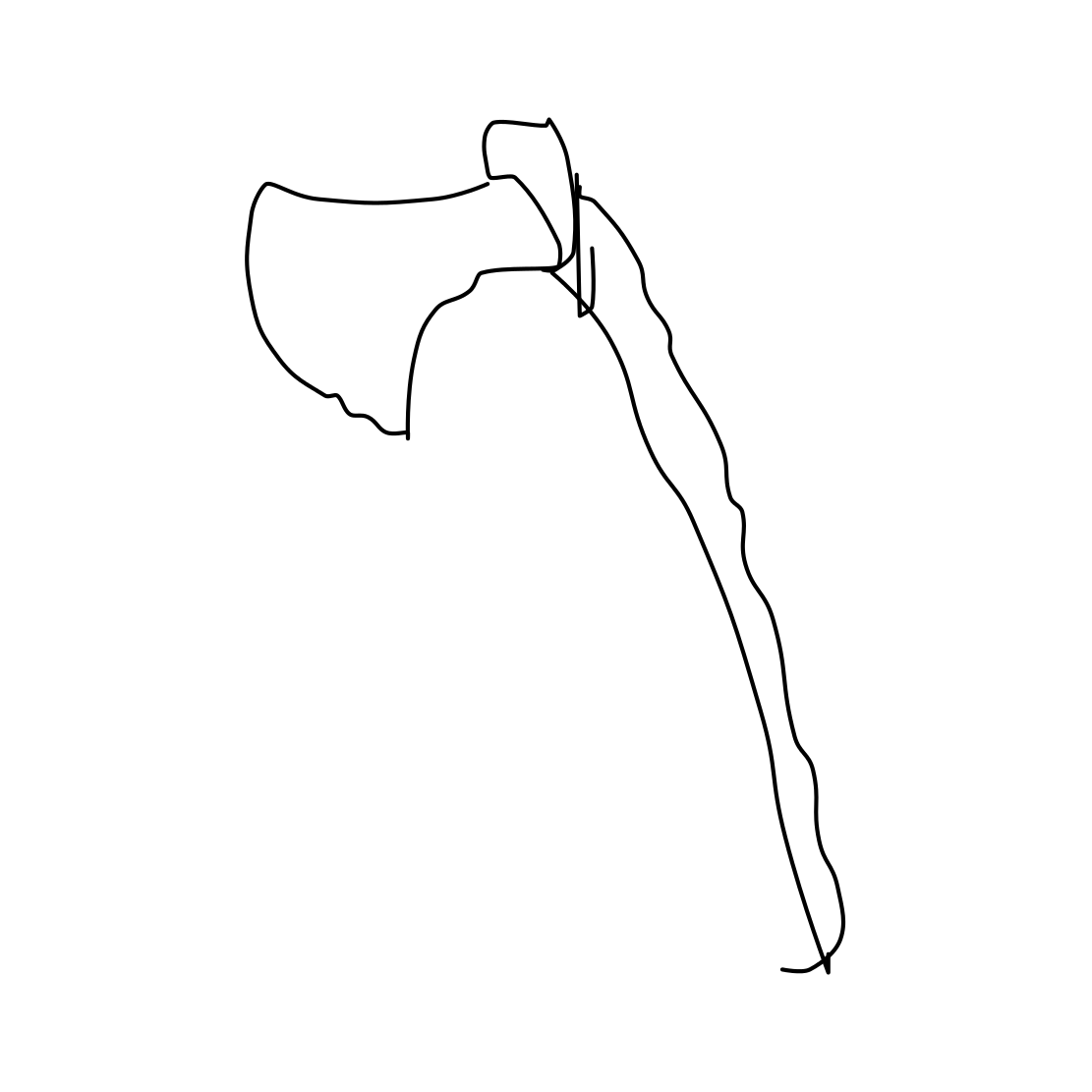}}
{\includegraphics[height=2.05cm]{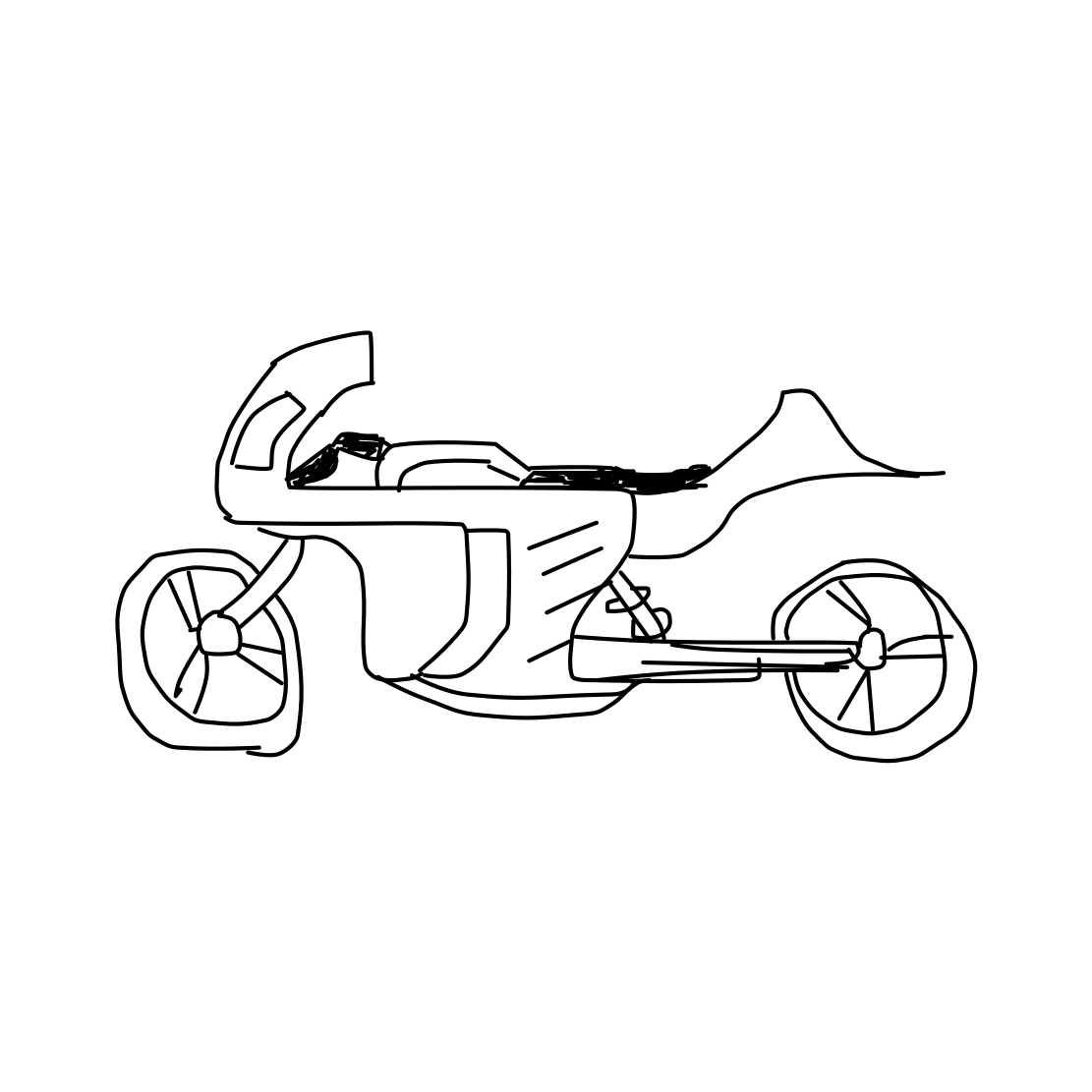}}
{\includegraphics[height=2.05cm]{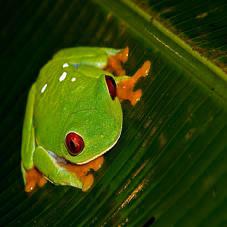}} 
 {\includegraphics[height=2.05cm]{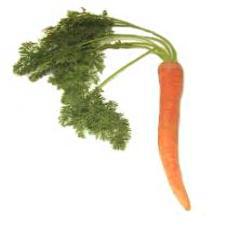}} \\
{\includegraphics[height=2.05cm]{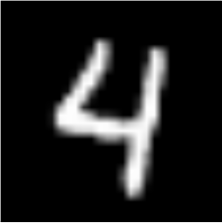}} 
{\includegraphics[height=2.05cm]{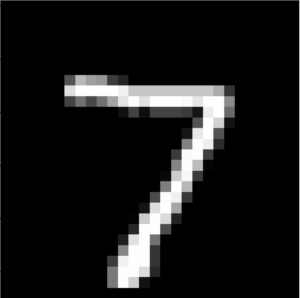}} \\
\caption{(From left to right, top to bottom) Two sample image instances of different classes from the chosen RSSCN, LSUN, TU-Berlin (sketch), TU-Berlin Extended and MNIST datasets, respectively.} %The samples of each dataset are placed column-wise.}
\label{fig:sampleimages}
\end{figure}

\noindent \textbf{Datasets:} For our experiments, we use three different vision datasets, each having different types of reported accuracies using various deep learning architectures, while apart from the pre-training part, the proposed GuCNet is not at all deep.  We want to demonstrate that in all cases, the proposed method can provide improvements by a considerable margin. For the first set of experiments, we use a small scale RSSCN aerial scene dataset \cite{zou2015deep}. This dataset contains 7 typical aerial scene categories. It has 400 image instances collected from the Google Earth under each semantic class, making it a total of 2800 images. Each image is of size of $400 \times 400$ pixels. The dataset is quite challenging due to its accumulation of different changing seasons, varying weathers, and different scales.  Since this dataset has only 7 classes, we use the standard MNIST dataset as the texture guide. The MNIST dataset \cite{lecun1998gradient} of digits comprises of 10 classes. This dataset is known to be highly discriminative as even a simple Imagenet pre-trained network is capable of yielding 99\% classification accuracy. We randomly choose 7 out of 10 classes as the texture data and associate them class-wise with the classes of the RSSCN dataset. Quite naturally, images of a particular class in these two datasets have absolutely no correlation (see Fig.~\ref{fig:sampleimages} for examples). For the prototype guide, we use a 128-dimensional 18-hot encoded vector as $g$, so that any two prototypes have a large Hamming distance of 36.

For the second set of experiments, we use a very large-scale indoor scene LSUN dataset \cite{yu2015lsun}. This dataset contains 10 million images for 10 scene categories. The varying image acquisition techniques make the image classification problem quite challenging. However, the current SOTA result is nearly 90\%.  Can we improve the accuracy further? Similar to the previous set of experiments, we use the well separable MNIST dataset with all 10 classes as the texture guide for classifying this data. For the prototype guide, we use a 128-dimensional 12-hot encoded vector as $g$ so that any two prototypes have a large Hamming distance of 24.

For our third set of experiments, we select a dataset consisting of a much larger number of classes. we choose the large-scale sketch database of TU-Berlin \cite{eitz2012humans}, which has 250 broad categories of semantic labels. Each class has 80 images, making it a total of 20,000 sketches. The selection of a proper texture guide for this data set is very difficult since there are very few image datasets that have above 250 semantic classes and yet perform as well as the MNIST dataset. We choose the extended version of the TU Berlin dataset which consists of the same 250 labels of natural (not sketches) images. It consists of a total of 204,489 natural images, distributed over 250 classes. We use two different experimental protocols for solving this problem: i) we put the images and sketches with the same semantic label in the same bin, and ii) we take two randomly chosen classes of images and sketches and put them into the same bin while training. The purpose of the experiment is to show that semantic labels of the guide data have no bearing on the performance of the proposed method. For the prototype guide, we use the first 250 one-hot encoded vectors of dimension 256. Fig \ref{fig:sampleimages} shows a few sample images of different semantic classes from each of the datasets that have been used in our experiments.

\begin{table}[!t]
\renewcommand{\arraystretch}{1.3}
% if using array.sty, 
 %\extrarowheight 
\caption{Classification performance of the proposed GuCNet architecture on RSSCN dataset. Here baseline for guide data (MNIST) is 99.80\%.}\label{tab:rsscn}
\centering
\begin{tabular*}{0.45\textwidth}{|l@{\extracolsep{\fill}}||c|}
%  \toprule
\hline
     \textbf{Model} & \textbf{ Accuracy(\%) \;}\\
    \hline 
    %color histogram &  &25 \\
    %TL~\cite{krizhevsky2012imagenet}& 73.20\%\\ 
    %LDA (SIFT)~\cite{xia2017aid} & 73.86\%\\ 
    %Zhou \etal~\cite{zou2015deep} & 77.00\%  \\ 
      LLC (CH)~\cite{xia2017aid} & 79.94\% \\ 
      SpLSA (SIFT)~\cite{xia2017aid} & 79.37\% \\ 
     VLAD (SIFT)~\cite{xia2017aid}& 79.34\% \\ 
      RGSIR~\cite{zafar2018novel} & 81.00\% \\ 
     %PIVWAH~\cite{ali2019modeling} & 82.48 \% \\
     AlexNet~\cite{zheng2019deep} &88.80\%  \\ 
      CaffeNet~\cite{zheng2019deep} &88.60\%  \\
      GoogleNet~\cite{zheng2019deep} &79.80\%  \\
      VGG-M~\cite{zheng2019deep} &87.30\%  \\
     VGG-VD16~\cite{zheng2019deep} &85.60\%  \\
      Conv5-MSP5-FV (vgg-s pretrain)~\cite{zheng2019deep}\;\;\;\; &95.40\%  \\ \hline
      \textbf{Baseline} & 88.39\% \\
     \textbf{GuCNet (Prototype)} & \textbf{97.36}\%  \\
     \textbf{GuCNet (Texture)} &  \textbf{99.11}\% \\
 %     \bottomrule
 \hline
\end{tabular*}
\end{table}

\begin{table}[!t]
\renewcommand{\arraystretch}{1.3}
% if using array.sty, 
% \extrarowheight 
\caption{Classification performance of GuCNet on LSUN dataset with the same guide data.}\label{tab:lsun}
\centering
\begin{tabular*}{0.45\textwidth}{|l@{\extracolsep{\fill}}||c|}
   \hline
     \textbf{Model} & \textbf{ Accuracy(\%) \;}\\
    \hline 
    %color histogram &  &25 \\
      Vanilla GAN~\cite{wang2019learning}& 70.50\% \\ 
     Labeled-samples~\cite{wang2019learning}& 77.00\% \\ 
     ds-cube~\cite{wang2017knowledge}& 83.00\%  \\
     Hybrid GAN~\cite{wang2019learning}& 83.20\% \\ 
     Normal BN-Inception + scene networks~\cite{wang2017knowledge} \;\; & 90.40\% \\ % + scene network
     Deeper BN-Inception + scene network~\cite{wang2017knowledge} \;& 90.90\% \\% + scene network
     SJTU-ReadSense~\cite{wang2017knowledge}& 90.40\%  \\
    Google~\cite{wang2017knowledge}& 91.20\% \\ 
    SIAT MMLAB~\cite{wang2017knowledge}& 91.60\% \\ \hline
     \textbf{Baseline} & 83.75\%\\
    \textbf{GuCNet (Prototype)}& \textbf{95.03\%} \\ 
    \textbf{GuCNet (Texture)} &  \textbf{94.86\%} \\
      \hline
\end{tabular*}
\end{table}

\noindent\textbf{Model architecture:} To learn $p_x$ and $p_y$ from the raw images, we use the commonly used ResNet-50 pre-trained network for extracting the features from all the data streams. This forms the baseline for our study. We keep the pre-trained network uniform in all our experiments to maintain uniformity and to avoid any network bias in the results. This yields us feature vectors of dimension 2048.  For extracting $p_x^l$ features, we sequentially use 3 layers of fully-connected networks (FCN) of dimensions $1024$, $512$, $K$, and $C$, where $C$ is the number of classes in that dataset. For the first two datasets $K=128$ as $C$ is either 7 or 10, while for the TU-Berlin sketch dataset $K=256$. Each layer, except the last layer, is followed by a drop-out layer with probability $0.5$. We use the $ReLU(\cdot)$ activation function after each layer of FCN, except the last layer, to induce non-linearity. % mention 4 million datasets only from lsun, extra layer after tu-berlin, 784 lsun feats, rsscn, clarify about f_x and (f_x  + fcn) 

\noindent\textbf{Training protocol and evaluation:} For pre-training the network, we used the stochastic gradient descent optimizer with a learning rate of 0.01 and a batch size of 64 samples. The TU-Berlin and the RSSCN datasets were trained for 20 epochs. The very large size of the LSUN dataset results in a satisfactory training performance within 2-3 epochs. For training both the variants of the GuCNet, we use the Adam optimizer with a learning rate of 0.001. We took a training batch size of 32 for the prototype-based guidance and a batch size of 64 (with 32 samples from $\mathcal{X}$ and 32 samples from $\mathcal{Y}$) for the texture-based guidance. We trained the proposed network for 50 epochs. We chose $\alpha=0.01$ heuristically, although we did not find much variation in results when $\alpha$ is varied around this value. A standard train:test split of 70:30 has been used in this study. It may be mentioned here that the guide data is needed only during the training process and one does not need it during testing.

%\hspace{2.5cm}

%\section{Results}\label{sec:results}
\noindent\textbf{Results:} We report the performance of the proposed GuCNet network and the other top performing frameworks in literature for a comprehensive comparative study. We report the performance of the RSSCN in Table~\ref{tab:rsscn}, LSUN in Table~\ref{tab:lsun}, and TU-Berlin in Table~\ref{tab:tub}. The current SOTA works in literature on the RSSCN dataset are by \cite{krizhevsky2012imagenet,xia2017aid,zou2015deep,zafar2018novel,ali2019modeling,zheng2019deep}. It can be observed from Table~\ref{tab:rsscn} that we outperform the SOTA~\cite{zheng2019deep} by a fairly good margin in both the variants of our network on the RSSCN dataset. One may argue that the part of the accuracy could be due to the pretraining of the RSSCN data using ResNet-50. We also show the performance of the baseline ResNet-50 trained data in the table, which is much lower than the GuCNet performance. For the LSUN dataset, we again outperform the SOTA~\cite{wang2017knowledge} by a considerable margin considering the fact that SOTA itself is quite high. Since the basic classifier network in the proposed work is quite a naive one, the key gain in the classification accuracy is derived from the ease of separability of the guide data which is about 99.8\% (see Table~\ref{tab:rsscn}) and not from the baseline which is only 83.75\%.

Likewise, we achieve superior results (an improvement of over 6\%) in comparison to contemporary techniques even for the TU-Berlin dataset when the baseline accuracy for the guide dataset is not as high as the MNIST dataset and the baseline accuracy for sketch data is only 69.90\%. Table~\ref{tab:tub} also shows that Prabhu \etal in \cite{prabhu2018hybrid} conducted experiments to quantify the performance of human subjects for the recognition of the image instances in the TU Berlin dataset. It was seen that even the humans could correctly classify the dataset with just 73.10\% accuracy. Thus, all these experiments highlight a significant boost in performance that can be achieved by the proposed GuCNet architecture with either variant of the guide data.

%\subsection{Ablation Study}
\noindent\textbf{Ablation Studies:}
In the first ablation experiment, we study if the choice of co-binning of classes of images from the guide set to the experimental data has any relevance. We put dissimilar classes of the images and sketches in the same bin by randomly shuffling the guide data and the experimental data subset correspondences. Table~\ref{tab:tub2} shows the difference in the performance of both the variants (same class binning as opposed to dissimilar class binning) while keeping the experimental protocols the same on the TU-Berlin dataset. We observe that the performance of the network has not been significantly affected, the marginal difference is possibly due to training difference. This clearly demonstrates that what semantic labels we give to the texture of the prototype guide may not affect the performance of the overall network. This is an important observation considering the amount of emphasis that the contemporaneous researchers are putting on the selection of highly descriptive semantic prototypes.

%\noindent{\textbf{Selection of $W$:}} 
To study the effect of the selection of the prototype vector $g$ for guided clustering, we perform our studies choosing four different types of prototype vector. In one case we use the previously defined same $g$ vector though a multi-hot encoded (18 for RSSCN and 12 for LSUN) $128$-dimensional vectors. This corresponds to choosing the maximum Hamming distance ($H_\text{max}$) possible for the chosen architecture. For the second experiment, we reduce $H_\text{max}$ by a factor of two, which reduces the separability among the prototypes. In the third case we further reduce the Hamming distance (H) to the minimum, \textit{i.e.,} $H=2$ that corresponds to just one-hot coded prototype. Quite naturally as $H$ increases, the separability increases and we expect a better performance as the mapping $\mathcal{F}_x : \mathcal{X} \mapsto \mathcal{Y}_c$ can be better realized. For the final set of experiments, we choose a random vector of length 128 (to emulate a prototype similar to word2vec) with values lying between [0,1). We report the results in Table~\ref{tab:w}. It is seen that an increased Hamming distance $H$ yields superior results and further it is much better than selecting word2vec representation. This is due to the fact that a randomly chosen prototype or a word2vec representation introduces highly variable inter prototype distances as opposed to the multi-hot encoded prototype. 

\begin{table}[!t]
\renewcommand{\arraystretch}{1.3}
% if using array.sty, 
 %\extrarowheight 
\caption{Performance comparison on TU-Berlin dataset for classification accuracy. Here baseline accuracy for guide data is 84.54\%.}\label{tab:tub}
\centering 
\begin{tabular*}{0.45\textwidth}{|l@{\extracolsep{\fill}}||c|}
 \hline
    \textbf{ Model} &  \textbf{Accuracy(\%) }\\
    \hline 
    AlexNet-SVM~\cite{krizhevsky2012imagenet} &67.10\% \\ 
    AlexNet-Sketch~\cite{krizhevsky2012imagenet} &68.60\% \\ 
    Sketch-A-Net SC~\cite{eitz2012humans}& 72.20\% \\
    Sketch-A-Net-Hybrid~\cite{eitz2012humans} \;\;\;\;\;\;\;\;\;\;\;\;\;\;\;\;\;\; &73.10\% \\
    ResNet18-Hybrid~\cite{he2016deep} &73.80\% \\
    Humans~\cite{prabhu2018hybrid} &73.10\% \\
    $\text{Sketch-A-Net-Hybrid}^2$ & 77.00\% \\
    $\text{Sketch-A-Net}^2$ & 77.00\% \\
    Alexnet-FC-GRU~\cite{prabhu2018hybrid} &79.95\% \\
    Zhang \etal~ \cite{zhang2019learning} & 82.95\% \\ \hline
    \textbf{Baseline} & 69.90\% \\
    \textbf{GuCNet (Prototype)}& \textbf{86.63\%} \\ 
     \textbf{GuCNet (Texture)}&  \textbf{89.26\%} \\
      \hline
\end{tabular*}
\end{table}

\begin{table}[!t]
\renewcommand{\arraystretch}{1.3}
% if using array.sty, 
 %\extrarowheight 
\caption{Ablation study showing the effect of different types of co-binning of texture classes from guide set.}\label{tab:tub2}
\centering
\begin{tabular*}{0.3\textwidth}{|l@{\extracolsep{\fill}}||c|}
%\begin{tabular}{|@{}l||c@{}|}
    \hline
    \textbf{ Dataset} (TU-Berlin) &  \textbf{Accuracy(\%)} \\
    \hline 
  \textbf{GuCNet (Texture):}&\\
 Same class binning \; &   89.26\%\\ 
 Dissimilar class binning \; &  90.05\% \\ 
\hline
\end{tabular*}
\end{table}

\begin{table}[!t]
\renewcommand{\arraystretch}{1.3}
% if using array.sty, 
% \extrarowheight 
\caption{Ablation study showing the effect of separability of prototypes in terms of Hamming distance $(H)$ on GuCNet performance.}\label{tab:w}
\centering
\begin{tabular*}{0.47\textwidth}{|l@{\extracolsep{\fill}}||c|c|c|c|}
%\begin{tabular}{||@{}l|@{}l|@{}l|@{}l||}
\hline
   \multirow{1}{*}{\textbf{Dataset}} & 
   \multicolumn{4}{c|}{\textbf{Separation of prototypes}}\\ \cline{2-5}
   & w2vec &$H=2$  & $\frac{H_\text{max}}{2}$& $H_\text{max}$\\
    \hline 
 \textbf{GuCNet (Prototype):} &&&&\\
 RSCCN dataset&  96.20\% &96.02\% &   96.27\%& 97.36\%\\ 
 LSUN dataset&   92.71\% & 94.60\% & 94.92\% &  95.03\%\\
\hline
\end{tabular*}
\end{table}

% \begin{table}[]
%     \centering
%     \begin{tabular}{c|c}
%         TU Berlin &  \\
%          TUB Sketches - 69.90 \\
% TUB Images - 84.54 \\
% LSUN & 83.75 \\
%     \end{tabular}
%     \caption{Caption}
%     \label{tab:my_label}
% \end{table}

\section{Conclusions}\label{sec:conclusion}
We propose a very simple yet novel, guided clustering-based neural network for obtaining an improved accuracy in classifying a difficult dataset. We leverage the ease of separability of a guide dataset to improve the separability of the mapped features for a challenging dataset while training a classifier. This helps in pushing the embeddings of the data instances far apart in the semantic feature space while making the embedding space further discriminative. It is the concept of using the help of a good guide data that makes our algorithm different from the existing architectures in literature. We demonstrated that one can use either texture-based or well separable prototype vector-based guides. We used a very simple network using only cross-entropy and difference losses to demonstrate the utility of the proposed concept. One may try to further improve the performance by using a more sophisticated network architecture. Currently, the assignment of a given prototype or a texture guide is arbitrary with respect to the semantics of a given class in the experimental data. The topology of the semantic classes in the learnt feature space implicitly depends on the feature mapping function. In this study, we have not taken into account the semantic topology of the texture or the prototype guides in the feature space while binning them with the experimental dataset. Ideally one would like to perform topology-preserving match making between the two datasets when the learning of the classifier is expected to be even better. Since the topologies of $\mathcal{X}_c$ and $\mathcal{Y}_c$ are not known apriori and are dependent of the mapping function itself, our future investigations would involve designing such a topology preserving network.% architecture.

\bibliographystyle{IEEEtran}

\bibliography{egbib}


%\section*{Acknowledgment}




%\bibliographystyle{IEEEtran}


% \end{thebibliography}

\end{document}